\def\paperID{5390}
\def\confName{CVPR}
\def\confYear{2024}

\def\paperTitle{MetaTra: Meta-Learning for Generalized Trajectory Prediction in Unseen Domain}

\def\authorBlock{    

 }

\newif\ifreview 
\newif\ifarxiv \newcommand{\arxiv}{\arxivtrue}
\newif\ifcamera 
\newif\ifrebuttal 

\arxiv 

\pdfoutput=1
\documentclass[10pt,twocolumn,letterpaper]{article}
\ifreview \usepackage[review]{cvpr} \fi
\ifarxiv \usepackage[pagenumbers]{cvpr} \fi
\ifrebuttal \usepackage[rebuttal]{cvpr} \fi
\ifcamera \usepackage{cvpr} \fi

\usepackage{graphicx}	
\usepackage{amsmath}	
\usepackage{amssymb}	
\usepackage{booktabs}
\usepackage{times}
\usepackage{microtype}
\usepackage{epsfig}
\usepackage[table,xcdraw,dvipsnames]{xcolor}
\usepackage{caption}
\usepackage{float}
\usepackage{placeins}
\usepackage{color, colortbl}
\usepackage{stfloats}
\usepackage{enumitem}
\usepackage{tabularx}
\usepackage{xstring}
\usepackage{multirow}
\usepackage{xspace}
\usepackage{url}
\usepackage{subcaption}
\usepackage{xcolor}
\usepackage[hang,flushmargin]{footmisc}

\ifcamera \usepackage[accsupp]{axessibility} \fi

% ...

\ifarxiv  \fi

\newcommand{\R}[1]{{%
    \textbf{%
        \ifstrequal{#1}{1}{\textcolor{red}{R#1}}{%
        \ifstrequal{#1}{2}{\textcolor{blue}{R#1}}{%
        \ifstrequal{#1}{3}{\textcolor{magenta}{R#1}}{%
        \ifstrequal{#1}{4}{\textcolor{teal}{R#1}}{%
                           \textcolor{cyan}{R#1}%
        }}}}%
    }%
}}

\usepackage{xr-hyper}

\makeatletter
\newcommand*{\addFileDependency}[1]{
  \typeout{(#1)}
  \@addtofilelist{#1}
  \IfFileExists{#1}{}{\typeout{No file #1.}}
}

\makeatother

\usepackage{algorithm} 
\usepackage{algpseudocode}  
\usepackage{amsmath} 

\definecolor{cvprblue}{rgb}{0.21,0.49,0.74}
\usepackage[pagebackref,breaklinks,colorlinks,citecolor=cvprblue]{hyperref}
\usepackage[capitalize]{cleveref}
\crefname{section}{Sec.}{Secs.}
\crefname{table}{Table}{Tables}
\crefname{figure}{Fig.}{Figs.}

\begin{document}
\title{\paperTitle}
\author{\authorBlock}
\maketitle
\begin{abstract}
Trajectory prediction has garnered widespread attention in different fields, such as autonomous driving and robotic navigation. However, due to the significant variations in trajectory patterns across different scenarios, models trained in known environments often falter in unseen ones. To learn a generalized model that can directly handle unseen domains without requiring any model updating, we propose a novel meta-learning-based trajectory prediction method called MetaTra. This approach incorporates a Dual Trajectory Transformer (Dual-TT), which enables a thorough exploration of the individual intention and the interactions within group motion patterns in diverse scenarios. Building on this, we propose a meta-learning framework to simulate the generalization process between source and target domains. Furthermore, to enhance the stability of our prediction outcomes, we propose a Serial and Parallel Training (SPT) strategy along with a feature augmentation method named MetaMix. Experimental results on several real-world datasets confirm that MetaTra not only surpasses other state-of-the-art methods but also exhibits plug-and-play capabilities, particularly in the realm of domain generalization.\cite{socialgan} \cite{evolvegraph} \cite{sophie} \cite{soccerdataset}
\end{abstract}
\vspace{-0.5cm}
\section{Introduction}
\indent Trajectory prediction plays a pivotal role in domains like robotic navigation, security surveillance, and sports analytics. Particularly with the rapid advancements of the autonomous driving industry, accurately predicting the trajectories of transportation elements is a key factor that influences the decision-making process of autonomous vehicles \cite{FEND,groupnet,HMM,HATN}.
\begin{figure}[htbp]
\vspace{-0.8cm}
\setlength{\abovecaptionskip}{0.1cm}
\setlength{\belowcaptionskip}{-0.2cm}
\centering
\includegraphics[width=0.47\textwidth]{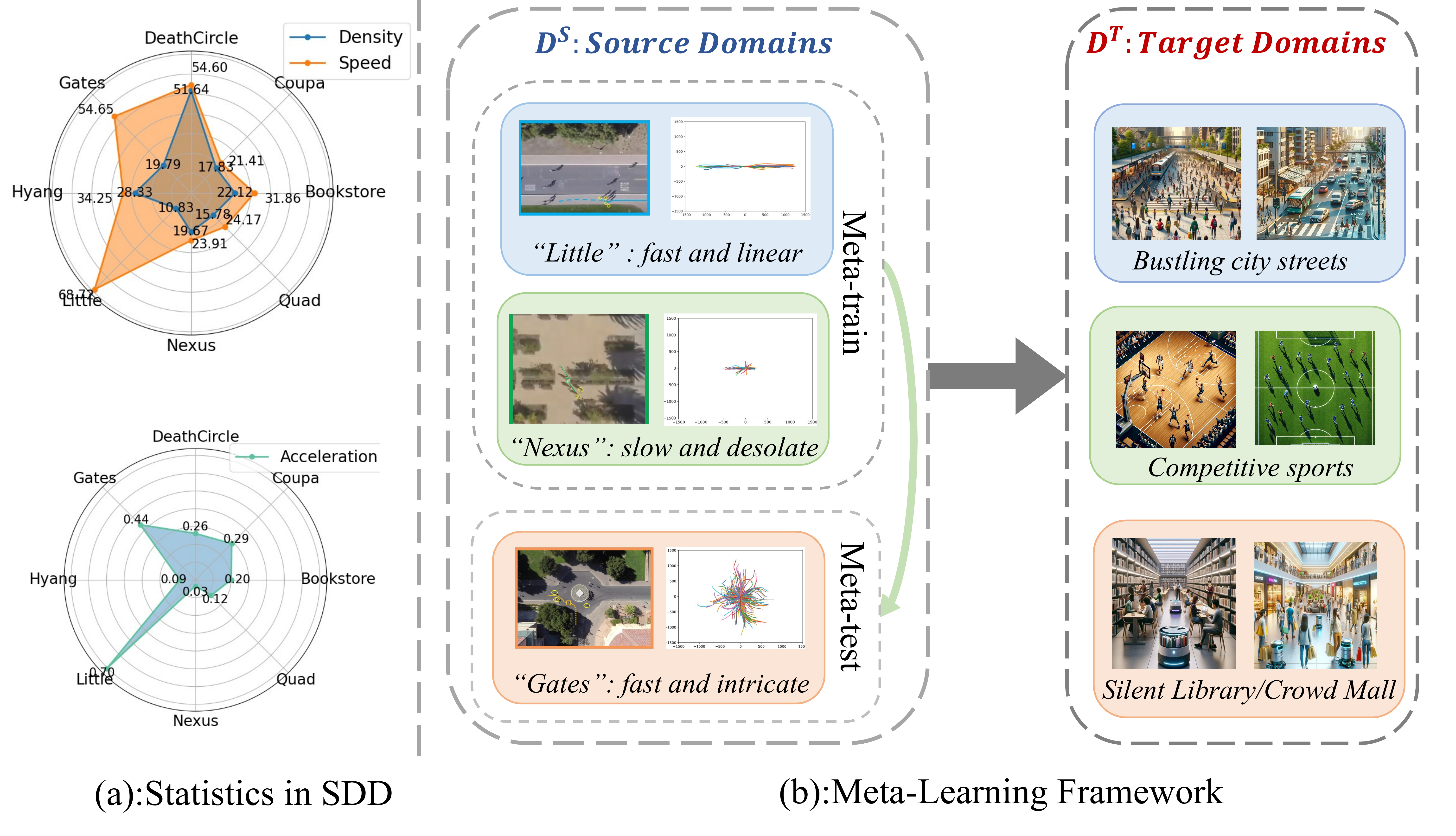} 
\caption{
    In part (a), we quantify the statistical biases in various scenarios within the SDD. The terms ``Density” ,``Speed”, and ``Acceleration” refer to the average pedestrian count, velocity, and acceleration metrics for each sequence, respectively. In part (b), our approach to the generalized trajectory prediction challenge is presented. It shows three source domains on the left for training, each with distinct trajectory distributions, and some unfamiliar target domains on the right for testing. Our model acquires domain-invariant universal knowledge by simulating domain shifts through meta-train and meta-test, thereby effectively generalizing to unseen domains. 
    } 
\label{fig:introduction} 
\vspace{-0.6cm}
\end{figure}
Initially reliant on physics-based techniques, such as the Kalman filter \cite{kalmanfilter} and Monte Carlo methods \cite{MonteCarlo}, the field of trajectory prediction has now shifted towards the adoption of machine learning approaches \cite{DynamicBayesianNetworks,HMM,SVM}. Further advancements have been realized with the integration of reinforcement learning frameworks, including MDP \cite{MDP} and GAIL \cite{GAIL}, as well as advanced deep learning algorithms like LSTM \cite{sociallstm} and Transformer \cite{transformer}. These developments have notably improved the accuracy in predicting track sequences. 
Simultaneously, the compatibility of graphs with Non-Euclidean data facilitates the effective modeling of complex interactions, prompting researchers to investigate the application of various graph structures in trajectory prediction \cite{GP-Graph,groupnet,HATN,Social-STGCNN}.
Also, generative models such as GAN \cite{GAN}, Diffusions \cite{diffusion1,diffusionleapfrog}, CVAEs \cite{CVAE}, and NPNS \cite{NPSN} contribute to capturing the randomness and uncertainty in real-world scenarios. 

However, current methods generally perform well within familiar environments but falter in new settings. This limitation primarily arises from an oversight of distribution discrepancies between training and test datasets, coupled with inadequate consideration of trajectory pattern variations across diverse scenarios. 
Our statistical analysis in Fig. \ref{fig:introduction}(a) highlights significant differences in density, speed, and acceleration across eight scenarios within the SDD dataset \cite{PecNet}, particularly in ``Gates", ``Little", and ``Nexus". As shown in Fig. \ref{fig:introduction}(b), we visualize the environment and all pedestrians' trajectories. 
When models are predominantly trained on scenarios such as ``Little", typified by rapid, linear pedestrian movement with distinct directional intent, they learn these patterns well. Similarly, in ``Nexus", where pedestrian density is low and movements are generally slower with fewer interactions, models may easily take effect due to the simplicity and consistency of the behavior.
However, this learned simplicity can be a hindrance when models encounter more complex settings.
If models are consummately trained on the predictable dynamics of ``Little" or the sparser interactions of ``Nexus" and then applied to the ``Gates" scenario, their predictive accuracy would likely diminish. 
This is due to insufficient optimization on the complex and erratic trajectory shifts observed in ``Gates", where the dense population of varying agents necessitates frequent and unpredictable path adjustments to evade collisions.

Addressing the aforementioned issue, several studies have employed Domain Adaptation (DA) \cite{ADAPT,alpaca-meta,tgnn} to bridge the gap between source and target domains by leveraging techniques that ensure feature consistency across domains, such as feature alignment and adversarial training. 
However, a limitation of DA is its inherent requirement for target domain data during the training phase. 
As illustrated in Fig. \ref{fig:introduction}(b), the deployed model should be able to generalize to unseen domains without any updating or fine-tuning. Additionally, some research has explored Domain Generalization (DG) \cite{domain-generation-good,domain-generation-survey}, aiming to utilize multiple source domains to cultivate models that can perform effectively in new, unseen environments. 
It typically encompasses strategies like data augmentation, advanced representation learning, and strategic learning policies. 
Nevertheless, the specific challenges associated with spatio-temporal aspects and labeling within trajectory prediction domains pose constraints on the application of prevalent DG strategies such as Explicit Feature Alignment \cite{CIDDG,DIVA}, Causality-Based Methods \cite{CTSDG}, Invariant Risk Minimization (IRM) \cite{IRM-Traj} and Ensemble Learning \cite{AdaRIP}. Meta-learning \cite{meta-survey,metamixup}, as a model-agnostic learning paradigm, is a novel alternative in this context. It has been widely applied across various challenges including few-shot learning, image generation, and domain generalization.

In this paper, we primarily investigate how to generalize trajectory prediction models to various domains, ensuring robust performance even in unseen domains. Initially, we delve deeply into two critical factors influencing trajectory prediction: the individual intentions and the interactive effects within group motional patterns. These elements endow the model with the capacity to handle complex scenarios. We propose a dual-path model \textbf{Dual} \textbf{T}rajectory \textbf{T}ransformer (Dual-TT) consisting of \textbf{I}nteracted-\textbf{T}emporal (IT) and \textbf{T}emporal-\textbf{I}nteracted (TI) pathways, amalgamating overarching trajectory features with interaction relationships. Building on this foundation, we propose a framework called \textbf{M}eta-Learning for \textbf{G}eneralized \textbf{T}rajectory \textbf{P}rediction (MGTP), adeptly simulating the transition between source and target domains through strategically reorganizing training data into meta-train and meta-test segments. This methodology facilitates the model's adaptation to source and target domains via implicit gradient alignment during the training phase. To circumvent the noise and instability arising from the prediction of uncertain trajectory points, we also design a \textbf{S}erial and \textbf{P}arallel \textbf{T}raining (SPT) strategy. This approach thoroughly computes gradients in both inner and outer loops to avoid converging to local optima. Concurrently, to increase the diversity of trajectory samples and thus reduce overfitting, we innovatively introduce the MetaMix strategy. In the unified feature space of meta-train and meta-test phases, we leverage probabilistic parameters from a variational encoder to sample multi-domain extra features, thereby enhancing trajectory prediction in the target domain. Our comprehensive methodology is named MetaTra: Meta-Learning for Generalized Trajectory Prediction in Unseen Domain.
The research contributions can be summarized as follows:
\begin{itemize}
\item To the best of our knowledge, our work is the first to apply a meta-learning framework to address the issue of generalization in trajectory prediction. The proposed Meta-Learning for Generalized Trajectory Prediction (MGTP) method entails a tailored delineation of tasks for the meta-train and meta-test phases. Concurrently, it incorporates a Serial and Parallel Training (SPT) strategy and the MetaMix method to augment the stability of the training process.

\item We propose a Dual Trajectory Transformer (Dual-TT) model, consisting of Interacted-Temporal (IT) and Temporal-Interacted (TI) pathways, designed to facilitate an integrative evaluation of the individual intention and the interactive effects within group motional patterns.

\item Experiments on three trajectory datasets confirm the consistent superiority of our method. We also show that the integration of MGTP can markedly promote the efficacy of various state-of-the-art prediction models.
\end{itemize}
\label{sec:intro}
\section{Related Work}
\ \indent \textbf{Trajectory Prediction}.
Recognizing the dynamic and complex nature of agent interactions in trajectory prediction, researchers have shifted towards various types of graphs \cite{GP-Graph,groupnet,HATN,sgcn} beyond grid-based representations. 
However, these methods typically utilize a single or sequential path for spatial and temporal modeling. For instance, Social-STGCNN \cite{Social-STGCNN} creates spatial graphs with GCNs before processing temporal dynamics with TCNs, potentially overlooking important global intent information. 
Models like STGAT \cite{stgat} and Social-BiGAT \cite{socialbigat} have enhanced graph-based approaches by incorporating attention mechanisms, using GAT to assign varying weights to different nodes. 
Nevertheless, such spatial graphs are typically homogeneous, failing to consider that different types of agents can produce distinct influences even at the same distance. 

\textbf{Domain Adaption (DA) and Domain Generalization (DG)}.
DA aligns distributions between training and testing datasets, with T-GNN \cite{tgnn} using a domain-adaptation loss function for this purpose.
Methods like HATN \cite{HATN} and OAMF \cite{MEKF} incorporate hierarchical structures and the MEKF algorithm for domain-specific adjustments. 
Ivanovic's work \cite{alpaca-meta} suggests ALPaCA for the final model layer, but DA's limitation is its reliance on target domain data. In contrast, DG trains on varied sources for better domain generalization without prior target data.
Techniques like data augmentation (rotations, flips, Mixup, CutMix \cite{mixup}) and adversarial representation learning (CIDDG \cite{CIDDG}, Diva \cite{DIVA}) enhance training diversity and create domain-agnostic features. 
AdaRIP \cite{AdaRIP} is an ensemble learning approach that combines predictions from different models, though it may increase computational demands.

\textbf{Meta Learning}.
Meta-learning, known as ``learning to learn", has gained traction with MAML \cite{MAML} for few-shot learning. Subsequent developments like FOMAML \cite{trainyourmaml} and Reptile \cite{Reptile} simplify second-order gradient computations. ANIL \cite{ANIL} shows that MAML's effectiveness comes from training approach and feature reuse rather than fine-tuning. Additionally, MLDG \cite{MLDG} expands meta-learning to domain generalization. M$^{3}$L \cite{M3L} and MetaMix \cite{metamixup} incorporate knowledge from meta-training to enhance diversity in meta-tests. Lastly, MVDG \cite{multi-view,MVDG} introduces multi-view regularization to reduce biases and local minima.

\label{sec:related}

\section{Method}
\label{sec:method}
\subsection{Problem Formulation}
\begin{figure*}[htbp]
\vspace{-0.77cm}
\setlength{\abovecaptionskip}{0.1cm}
\setlength{\belowcaptionskip}{-0.2cm}
    \centering
    \includegraphics[width=0.9\textwidth]{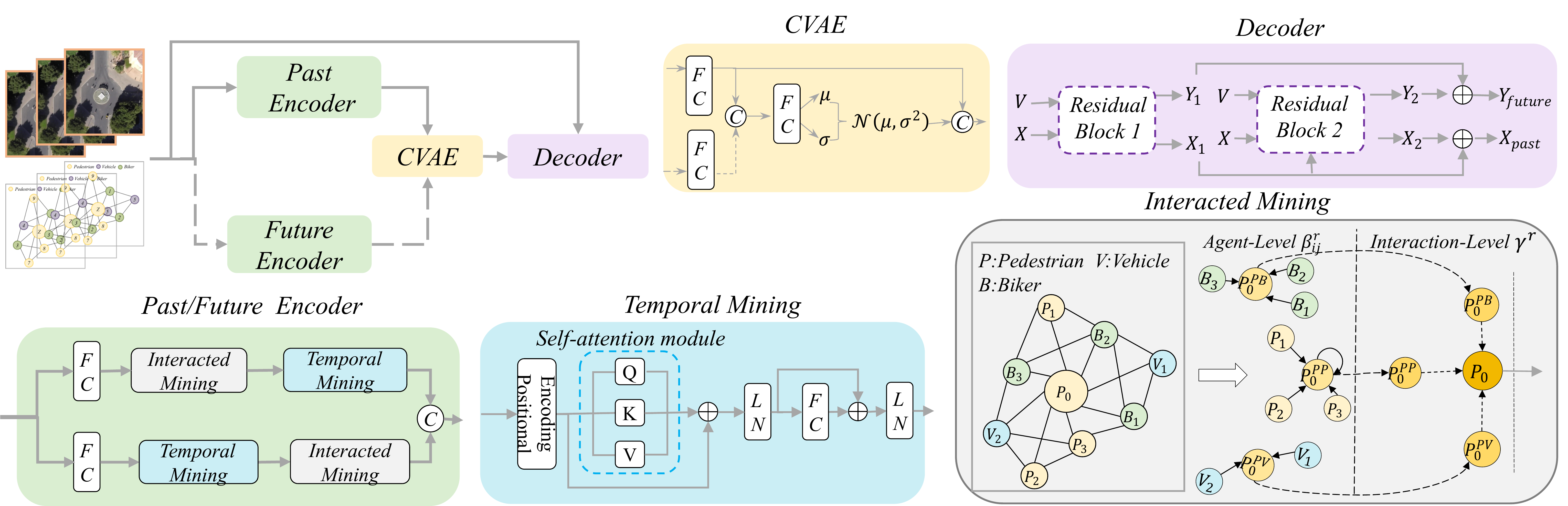} 
    \caption{Model architecture of the proposed Dual Trajectory Transformer (Dual-TT).} 
    \label{fig:backbone} 
\vspace{-0.5cm}
\end{figure*}
\ \indent Assuming the presence of $N$ agents concurrently moving in the scene, we can observe the motion trajectory, denoted as $X^{i}$, of the $i$-th agent over the past $t_{obs}$ moments. The positional coordinates of this trajectory are represented by $X^{i} = \{o^{i}_{1},\ldots, o^{i}_{t_{obs}}\}$. The orientation of this work is to predict the future trajectory up to time $t_{pre}$, denoted as $Y^{i} = \{o^{i}_{t_{obs}+1},\ldots, o^{i}_{t_{pre}}\}$. Here, $o^{i}$ represents the spatial coordinates in either two-dimensional or three-dimensional space.

Particularly, we have access to a dataset comprising $N_{S}$ source domains $D^{S} = \{D^{1},\ldots,D^{N_{S}}\}$, within which the trajectory prediction model is trained. Concurrently, there is an aspiration for the model to be directly applicable to a target domain $D^{T}$ emanating from novel environments.
Our aim is to develop a predictive model $f(\cdot|\theta)$ endowed with generalizability parameters. For trajectories deriving from unseen domains, the predictive outcome still closely approximates the true values.
%
\subsection{Dual Trajectory Transformer}
\ \indent A generic trajectory prediction model capable of adapting to various scenarios necessitates a comprehensive consideration of multifaceted factors. Precision in predicting the uncertain future trajectories of multi-agents requires a holistic consideration of each agent's intentions and destinations, which can be approximately inferred from their historically observed trajectories. Additionally, it is essential to consider the impact of various inter-agent relationships on the collective's overall trajectory patterns. For instance, multiple pedestrians might proceed abreast in the same direction on sidewalks, while athletes on a sports field engage in strategic positioning based on their offensive or defensive roles. In summary, we introduce the Dual Trajectory Transformer (Dual-TT) model (see Fig. \ref{fig:backbone}), which encompasses both Temporal and Interacted Mining components to handle complex scenarios effectively.

\textbf{Temporal Mining.}
Trajectory prediction pertains to a sequence prediction problem, where the observed trajectories of agents inform the prognostication of their prospective movement patterns. We employ the classic transformer module to capture the contextual attention in trajectory sequences. This is primarily attributable to the dynamic properties of trajectory data, wherein the prediction of future movement necessitates a simultaneous consideration of both an exhaustive historical representation and their localized patterns. For the initial setup, we offset the input trajectory $X$ with the averaged coordinates of the final observation point, calculating the relative position $o'^{i}_{t} = o^{i}_{t}-\frac{1}{N}\sum_{i=1}^{N}o^{i}_{t_{obs}}$ to mitigate the disturbance of scene sizes and absolute coordinate values. Subsequently, the fully connected layer is followed to attain the embedding $h^{i}$. The self-attention module sequentially projects the features into a query matrix $Q$, a key matrix $K$, and a value matrix $V$. Thereafter, the dot products between the queries and keys are computed to determine the attention values at various moments, and a weighted sum operator is adopted to aggregate the final temporal feature value $h^{i}_{\mathcal{T}}$. This entire process is repeatedly executed with multi-head attention mechanism $k$ times and the outcomes are synthesized.

\textbf{Interacted Mining.}
Within the group, agents interact through multiple relations, with their trajectories giving rise to diverse collective movement patterns, such as the deflection in the travel direction of pedestrians when they are obstructed. Manifestly, the type and magnitude of these interactions necessitate thorough excavation. Some studies \cite{STAR} have simplified the interactions among all agents yet overlooked the more comprehensive information and rich semantics contained therein.

Inspired by the research of heterogeneous graph neural networks \cite{heterogeneous}, we contemplate the influences each node exerts within the network of disparate interactive relationships. Subsequently, we aggregate the various influential forces impinging upon an agent. For instance, in an NBA match, a player's movement is predicated not only on giving-and-going with teammates but also on eluding opponents. First of all, we delineate a scenario having $R$ relationship types. Next, we employ the graph attention mechanism to compute the weights of influence between nodes within a specified spatial distance. Given a pair of agents with a relationship $r$, the significance of their mutual influence is computed by the softmax function as follows:
\begin{equation}
	\setlength{\abovedisplayskip}{3pt}
	\setlength{\belowdisplayskip}{3pt}
	\beta^{r}_{ij} = \frac{exp(a_{r}^{T}\cdot[h^{i}||h^{j}])}{\sum_{k\in \mathcal{N}_{i}^{r}}exp(a_{r}^{T}\cdot[h^{i}||h^{k}])},
\end{equation} 
Here, $\mathcal{N}_{i}^{r}$ represents the neighbors of node $i$ on the meta-relation $r$. $h^{i}$, $h^{j}$ and $h^{k}$ denote the projected features of the nodes, while $a_{r}$ constitutes the attention vector pertinent to the designated interaction, which is concurrently optimized during the optimization phase. Notably, the influence ratio $\beta^{r}_{ij}$ is inherently asymmetric, endowing the inter-agent influences with unidirectional interactions.
To thoroughly consider the complexity of relationships, we utilize K-head attention to concatenate learned trajectory features. The semantic features above are ultimately achieved through the aggregation of neighboring nodes.
\begin{equation}
	\setlength{\abovedisplayskip}{5pt}
	\setlength{\belowdisplayskip}{5pt}
	{\hat{h}_{r}^{i}} =
	\underset{r=1}{\overset{R}{\Vert}}\sigma(\sum_{j\in\mathcal{N}_{i}^{r}} \beta^{r}_{ij} \cdot h^{j}).
\end{equation}  
Ultimately, we can obtain $R$ relation-specific node embeddings for one agent. During the collective motion process, agents often exhibit varying attention towards different relationships. We employ a relational attention vector $q$, which measures the similarity between each interaction component, to determine the interaction weights. We further normalize these attention weights across all interactions using the softmax function as follows:
\begin{equation}
	\setlength{\abovedisplayskip}{5pt}
	\setlength{\belowdisplayskip}{5pt}
	\gamma^{r} = \frac{exp(\frac{1}{N}\sum_{i\in\mathcal{V}}q^{T}\cdot tanh(W\cdot{\hat{h}_{r}^{i}}+b))}{\sum_{r=1}^{R}exp(\frac{1}{N}\sum_{i\in\mathcal{V}}q^{T}\cdot tanh(W\cdot{\hat{h}_{r}^{i}}+b))}.
\end{equation} 
The interaction effects of neighbors can be aggregated using their coefficients as weights for feature fusion:
\begin{equation}
	\setlength{\abovedisplayskip}{3pt}
	\setlength{\belowdisplayskip}{3pt}
	h_{\mathcal{R}}^{i} = \sum\limits_{r=1}^{R}\gamma^{r}\cdot{\hat{h}_{r}^{i}}.
\end{equation}

\textbf{Model Framework and Optimization.}
The aforementioned temporal mining module solely considers the intrinsic patterns of trajectories, lacking an induction of group influence. Conversely, interacted mining mainly focuses on the mutual interactions among nearby agents, yet it tends to overlook the dynamic evolution of trajectories and the agents' individual intents. To synthesize the strengths of these components, we leverage both sets of features within a dual-path model framework. In the temporal-interacted branch, a temporal mining encoder takes the trajectory coordinates of group members at moments $\{1,\ldots,T\}$ as input and then executes interacted mining at time $T$ to yield intermediary features. This process is independently and thoroughly analyzed for each trajectory, culminating in the amalgamation of the comprehensive features of neighboring agents. In the interacted-temporal branch, the influence features between groups at each time are computed separately to explore the interactive impact at that moment. Temporal mining is subsequently applied to obtain the coefficients of these interactive impacts. The concatenation $[h^{i}_{\mathcal{T}}\Vert h_{\mathcal{R}}^{i}]$ results in the fused feature embedding for node $i$ (see Fig. \ref{fig:backbone}).

In the prediction phase, we utilize the conditional variational autoencoder (CVAE) model to achieve trajectory probability prediction with a reasonable degree of randomness. Using the observed historical trajectories $X$, we calculate the conditional probability $P(Y|X)$ for predicting future trajectories $\hat{Y}$. Additionally, we introduce a latent vector $Z\in\mathcal{R}^{N*d}$ where $N$ and $d$ represent the number and dimension of samples, subsequently compute the posterior probability distribution of future trajectories.
\begin{equation}
	\setlength{\abovedisplayskip}{3pt}
	\setlength{\belowdisplayskip}{3pt}
	p(Y|X) = \int p(Y|X,Z)p(Z|X)dZ.
\end{equation}
Here, $p(Z|X)$ follows a Gaussian distribution while $p(Y|X,Z)$ is the conditional probability distribution based on $X$ and $Z$. In order to solve the above equation, we use the evidence lower bound (ELBO) to calculate the loss $L_{pred}$ based on the maximum likelihood estimation method.
\begin{equation}
	\setlength{\abovedisplayskip}{3pt}
	\setlength{\belowdisplayskip}{3pt}
 L_{pred} = -\mathbb{E}_{q(Z|X,Y)}p(Y|X) + KL(q(Z|X,Y)||p(Z|X)).
\end{equation}
Inspired by Hypertron \cite{hypertron}, we utilize a past encoder to model $p(Z|X)$ based on historical trajectories, while also using a future decoder $q(Z|X,Y)$ to model both past and future trajectories. As for $q(Y|X)$, we use residual decoders which both reconstruct past sequences and predict future sequences. 
During the training phase, the past and future trajectories are separately encoded and concatenated using the proposed Dual-TT module. Two separate perceptrons are used to obtain the parameters $\mu$ and $\sigma$ for sampling $q(Z|X,Y)=\mathcal{N}(\mu_{q},\sigma_{q})$, which is then acquired by the Gumbel Softmax. During the testing phase, $Z$ is only sampled by $p(Z|X)=\mathcal{N}(\mu_{p},\sigma_{p})$. Afterwards, $Z$ is concatenated with the past trajectory features and input together with the historical trajectory $X$ into the residual decoders \cite{spectral}, which contain two GRU-MLP encoder-decoder modules, to obtain the reconstructed and predicted trajectories. 
We define our overall loss by integrating the evidence lower bound with the deviation between the predicted reconstructed trajectories $\hat{X}$, future trajectories $\hat{Y}$, and their actual real-world counterparts $X$ and $Y$. The formulation is as follows:
\begin{small} 
\begin{equation}
	\setlength{\abovedisplayskip}{5pt}
	\setlength{\belowdisplayskip}{3pt}
L=\Vert\hat{Y}-Y\Vert+\zeta\cdot KL(q(Z|X,Y)||p(Z|X))+\eta\cdot\Vert\hat{X}-X\Vert.
\end{equation}
\end{small}

\textbf{Compared to Previous Work.}
Our methodology aligns with prior studies like STAR \cite{STAR} and STT \cite{many} in employing the attention-based transformer mechanism. Yet, we innovate by implementing symmetrical dual structures for feature fusion. Similar to SGCN \cite{sgcn}, our model extracts trajectories from both temporal and spatial perspectives. However, we have developed a more detailed characterization of diverse interaction semantics, complemented by the integration of a more sophisticated decoder.
\subsection{Generalized Trajectory Prediction}
\begin{figure*}[htbp]
\vspace{-0.77cm}
\setlength{\abovecaptionskip}{0.1cm}
\setlength{\belowcaptionskip}{-0.2cm}
    \centering
    \includegraphics[width=0.9\textwidth]{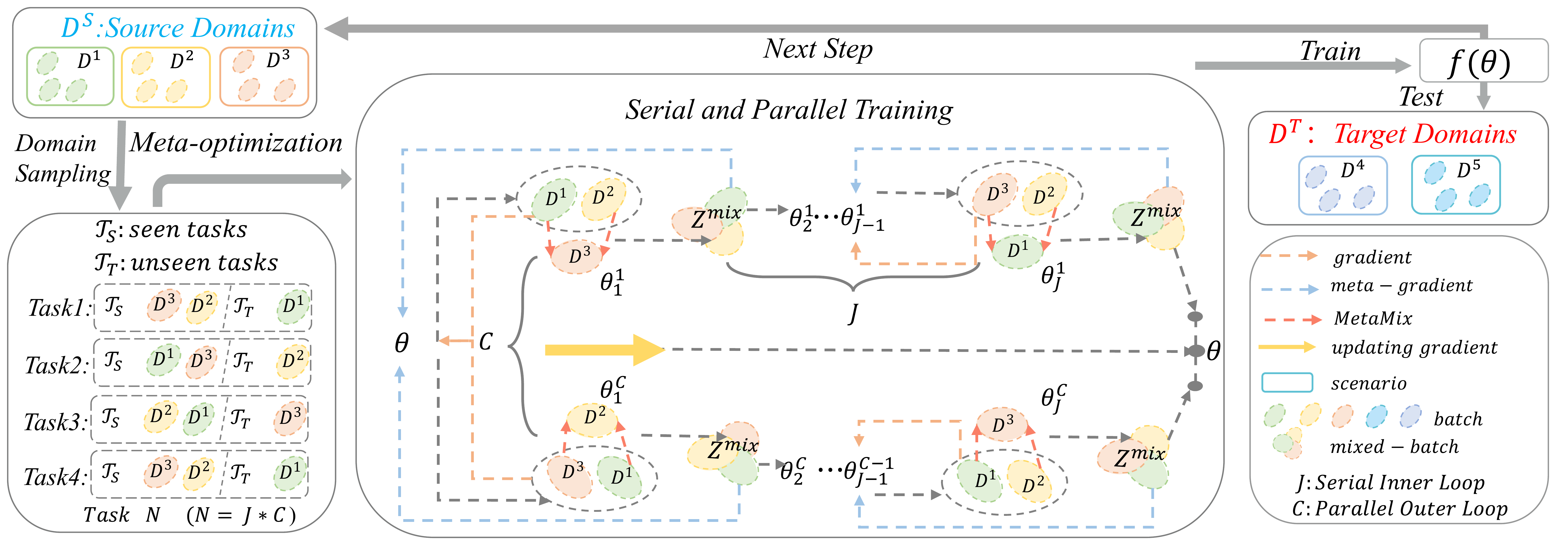} 
    \caption{Meta-learning framework for Generalized Trajectory Prediction (MGTP).} 
    \label{fig:meta-learning framework} 
\vspace{-0.5cm}
\end{figure*}
\ \indent We now turn our attention to the objective of training a trajectory prediction model $f(\cdot|\theta)$ on the samples from the source domain $D^{S}$, and then achieve generalization on the target domain $D^{T}$. Intuitively, this objective can be achieved by optimizing the model parameters using losses $L_{S}$ and $L_{T}$ derived from the above source and target domains.
\begin{equation}
	\setlength{\abovedisplayskip}{3pt}
	\setlength{\belowdisplayskip}{3pt}
	\sum_{(x^{i},y^{i})\in D^{S}}L_{S}(x^{i},y^{i};\theta)+\sum_{(x^{j},y^{j})\in D^{T}}L_{T}(x^{j},y^{j};\theta).
\end{equation}
Based on this, we introduce the Meta-Learning for Generalized Trajectory Prediction (MGTP) framework (see Fig. \ref{fig:meta-learning framework}), which primarily comprises the following components.

\textbf{Meta-Learning for Unseen Domain.} 
To facilitate trajectory prediction in unseen domains, it is crucial that the feature distribution demonstrates substantial generalizability. Meta-learning is commonly utilized to optimize the learning process, mainly when training samples are limited. A classic paradigm within meta-learning involves adopting an initialization-differentiation methodology across multiple tasks, exemplified by the Model-Agnostic Meta-Learning (MAML).  This approach involves iterative gradient updates—both inner and outer loops—to derive initial parameters that can be rapidly fine-tuned across various domains. As an advancement, Reptile employs first-order gradient updates to enhance the efficiency of the learning process. 

Given the unknown feature space of trajectories in the test phase, we employ a cyclic two-order meta-learning approach to replicate the generalization process. As detailed previously, during the meta-training phase, we select $\vert D^{S}\vert$-1 domains as source domains for a first-order parameter update. Using the loss defined earlier, the temporary parameters $\theta'$ can be represented as $\theta'=\theta - \lambda_{S}\nabla_{\theta} L_{S}(X,Y;\theta)$. Subsequently, in the meta-testing phase, we compute the loss $L_{T}(X,Y;\theta')$ using $\theta'$ on the target domain to efficaciously enhance the model's generalization capability. The final step involves updating the original parameters $\theta$ using a combined loss. This methodology allows to optimize for generalization, where the update process can be concisely delineated as follows.
\begin{equation}
		\setlength{\abovedisplayskip}{3pt}
	\setlength{\belowdisplayskip}{3pt}
	\theta = \theta - (\lambda_{S}\nabla_{\theta} L_{S}(X^{S},Y^{S};\theta)+\lambda_{T}\nabla_{\theta}L_{T}(X^{T},Y^{T};\theta'))
\end{equation}
The amalgamation of gradients from both directions fosters a trajectory predictor, trained across source and target domains, to derive mutual benefits.

\textbf{Serial and Parallel Training.}
Through synergistic gradient optimization across source training domains, the proposed meta-learning module accomplishes model generalization in out-of-domain contexts. However, intuitively, trajectory prediction diverges from classification tasks in that its solution space within various scenes cannot be fully delineated. Gradient updates confined to a single target domain are prone to introducing noise and adversely impacting model stability. Additionally, sparse cross-domain training may make the model ensnare in the local minimum.
We propose a training strategy that combines serial and parallel training routes (called Serial and Parallel Training) to address these challenges. In the inner loop, isolated gradient updates do not guarantee comprehensive parameter optimization and may culminate in a suboptimal local minimum. Drawing on the Reptile training methodology, we aim for the trajectory predictor $f(\cdot|\theta)$ to converge towards a solution proximal to each task’s manifold of optimal solutions. Consequently, we devise a serial inner loop, repeatedly samples seen tasks $\tau_{S}$ and unseen tasks $\tau_{T}$ within the source domain $D^{S}$. Using gradient descent for parameter updates, we define the parameter updating operation as follows.
\begin{equation}
		\setlength{\abovedisplayskip}{3pt}
	\setlength{\belowdisplayskip}{3pt}
U_{i}(\theta_{i}) = \theta_{i-1}-\lambda(L_{\tau_{S}^{i}}(\theta_{i-1})+L_{\tau_{T}^{i}}(\theta_{i-1}-\lambda(L_{\tau_{S}^{i}}(\theta_{i-1})))).
\end{equation}
Upon completion of $J$ serial updates through the operation $U_{J}$, the optimized parameters $\widetilde{\theta}_{J}$  for this serial trajectory are obtained. Regarding the original parameters $\theta_{0}$, the updating is executed using $\theta_{0} = \theta_{0} + \kappa(\widetilde{\theta}_{J} - \theta_{0})$.

Similarly, for the outer loop, if one were to rely solely on single-view sampling within the outer loop, this would represent merely a transient observational state, highly susceptible to noise and instability. Previous methodologies, such as MVDG \cite{MVDG}, have highlighted that optimization in a single direction could lead to inaccuracies, and the resulting sharp minima may compromise generalizability. Inspired by them, we adopt a parallel outer loop approach to comprehensively explore the optimal position within the parameter space. In each iteration, multiple parameter sets $\{\widetilde{\theta}_{J}^{1},\ldots,\widetilde{\theta}_{J}^{C}\}$ are obtained. Ultimately, this paper employs an averaging method for parallel updating of parameter values:
\begin{equation}
		\setlength{\abovedisplayskip}{3pt}
	\setlength{\belowdisplayskip}{3pt}
\theta=\theta+\kappa(\sum_{c=1}^{C}\frac{1}{C}\widetilde{\theta}_{J}^{c}-\theta).
\end{equation}

\textbf{Domain Feature Augmentation.}
Our meta-learning framework emulates the generalization process by performing parameter updates across source and target domains, thereby getting a trajectory predictor applicable in the unseen domain. During this process, as the meta-test phase utilizes tasks from only a single test domain, effectively enhancing the diversity of its samples is conducive to augmenting the model's generalization capability \cite{open}. However, gathering sufficient target-domain training samples is often impractical. Given that the source domain encompasses a more varied and optimized distribution of feature information, we propose MetaMix, a method to enrich target domain trajectory samples using source domain data. As depicted in Fig. \ref{fig:meta-learning framework}, during the meta-test phase, we implement a mixup procedure combining trajectory features from the test domain with those from the source domain. These inter-domain features assist in enhancing the diversity of the features.

Delving specifically into the proposed model, we utilize a gaussian prior distribution during the training process to sample latent embedding $Z$ within mean $\mu$ and variation $\delta$. In the meta-training process, $P(Z|X^{S},Y^{S})$ represents the integrated high-level features originating from the source domain, as opposed to specific sample features. Naturally, in the meta-testing phase, MetaMix aims to resample extra features from this distribution and inject them into the target dataset.
For each batch of $\mathcal{B}$ trajectories from the test domain, with latent features $\{Z^{T}_{i}\}_{i=1}^{\mathcal{B}}$ obtained from variational encoder, we aim to resample corresponding additional features $\{Z^{add}_{i}\}_{i=1}^{\mathcal{B}}$ from the source domain, resulting in the mixed features $\{Z^{mix}_{i}\}_{i=1}^{\mathcal{B}}$. The feature fusion process initiates by sampling a mixing coefficient $\rho\sim Beta(1,1)$, and simultaneously generating $\{Z^{add}_{i}\}_{i=1}^{\mathcal{B}}$ from a normal distribution $\mathcal{N}(\mu^{S},\delta^{S})$. The following formula can encapsulate the calculation process:
\begin{equation}
		\setlength{\abovedisplayskip}{3pt}
	\setlength{\belowdisplayskip}{3pt}
	Z^{mix} = Z^{T} + (1-\rho)Z^{add}.
\end{equation}
During the meta-testing phase, we concatenate the mixup-enhanced latent embeddings $Z^{mix}$ with the initial input $Z^{T}$ of the decoder, culminating in $2\mathcal{B}$ predicted trajectories as outputted by the decoder. Concurrently, we replicate the true past and future samples from the test tasks, which are then correspondingly employed to compute the test loss.
\vspace{-0.3cm}
\section{Experiments and Analysis}
\begin{table*}[htpb]
\vspace{-0.3cm}
\setlength{\abovecaptionskip}{0cm}
\setlength{\belowcaptionskip}{-0.2cm}
\centering
\caption{\text{minADE$_{20}$}/\text{minFDE$_{20}$} (meters) results on ETH-UCY. Agent Former performs best in previous methods (marked with $*$) and gets better when utilizing MGTP as a plugin. Our Dual-TT, integrated with MGTP, achieves the best results. (bold: best, underline: runner-up)}
\label{tab:ETH-UCY}
\resizebox{0.97\textwidth}{!}
{
\begin{tabular}{l|c c c c c|c c|c c}
\hline \hline
\multirow{2}{*}{Subset} & Social-STGCNN & Trajectron++ & PECNet & STAR & GroupNet & Agent Former$^*$ & Agent Former$^*$ & Dual-TT & Dual-TT\\
& \textcolor{blue}{CVPR20\cite{Social-STGCNN}} & \textcolor{blue}{ECCV20\cite{Trajectron++}} & \textcolor{blue}{ECCV20\cite{PecNet}} & \textcolor{blue}{ECCV20\cite{STAR}} & \textcolor{blue}{CVPR22\cite{groupnet}} & \textcolor{blue}{ICCV21\cite{AgentFormer}} & +MGTP &  & + MGTP\\
\hline
ETH & 0.64/1.11 & 0.61/1.02 & 0.54/0.87 & 0.36/0.65 & 0.46/0.73 & 0.45/0.75 & 0.36/0.57 & \underline{0.29/0.54} & \textbf{0.24/0.46} \\
HOTEL & 0.49/0.85 & 0.19/0.28 & 0.18/0.24 & 0.21/0.41 & 0.15/0.25 & 0.14/0.22 & \textbf{0.13/0.21} & 0.15/0.28 & \underline{0.13/0.23} \\
UNIV & 0.44/0.79 & 0.30/0.54 & 0.35/0.60 & 0.33/0.67 & 0.26/0.45 & \underline{0.25/0.45} & \textbf{0.24/0.43} & 0.31/0.61 & 0.29/0.58 \\
ZARA1 & 0.34/0.53 & 0.24/0.42 & 0.22/0.39 & 0.29/0.52 & 0.21/0.39 & \textbf{0.18/0.30} & 0.19/0.30 & 0.23/0.46 & \underline{0.18/0.39} \\
ZARA2 & 0.30/0.48 & 0.18/0.32 & 0.17/0.30 & 0.22/0.46 & 0.17/0.33 & \underline{0.14/0.24} & \textbf{0.13/0.22} & 0.19/0.38 & 0.16/0.33 \\
\hline
AVG & 0.44/0.75 & 0.30/0.51 & 0.29/0.48 & 0.28/0.54 & 0.25/0.44 & 0.23/0.39 & \underline{0.21/0.35} & 0.23/0.45 & \textbf{0.20/0.40} \\
\hline
\end{tabular}
}
\vspace{-0.2cm}
\end{table*}

\begin{table*}[htpb]
\setlength{\abovecaptionskip}{0cm}
\setlength{\belowcaptionskip}{-0.2cm}
\centering
\caption{\text{minADE$_{20}$}/\text{minFDE$_{20}$} (pixels) results on SDD. PECNet performs best in previous methods (marked with $*$) and gets better when utilizing MGTP as a plugin. Our Dual-TT, integrated with MGTP, achieves the best results. (bold: best, underline: runner-up)}
\label{tab:SDD}
\resizebox{0.98\textwidth}{!}
{
\begin{tabular}{l|c c c c c |c c|c c}
\hline \hline
\multirow{2}{*}{agents type} & Social-STGCNN & Trajectron++ & STAR  & Agent Former & Group Net & PECNet$^*$ & PECNet$^*$ & Dual-TT & Dual-TT\\
& \textcolor{blue}{CVPR20\cite{Social-STGCNN}} & \textcolor{blue}{ECCV20\cite{Trajectron++}} & \textcolor{blue}{ECCV20\cite{STAR}} & \textcolor{blue}{ICCV21\cite{AgentFormer}} & \textcolor{blue}{CVPR22\cite{groupnet}} & \textcolor{blue}{ECCV20\cite{PecNet}} & + MGTP &  & +MGTP\\
\hline
pedestrian & 20.25/35.75  & 19.30/32.70 & 17.63/36.21 & 10.32/18.30
& 9.31/16.11  & 9.96/15.88  & \underline{8.91/13.38} & 12.79/15.88 & \textbf{8.49/11.85}  \\
all-agents     & 27.23/41.44
&  25.66/39.76  & 27.73/52.37 & 16.84/29.82 & 14.05/27.67 & 14.18/24.47 & \underline{10.97/18.40} & 16.27/29.38 & \textbf{10.13/17.05} \\
\hline
\end{tabular}
}
\vspace{-0.6cm}
\end{table*}

\begin{figure*}[htbp]
\vspace{-0.6cm}
\setlength{\abovecaptionskip}{0.3cm}
\setlength{\belowcaptionskip}{-0.2cm}
    \centering
    \begin{subfigure}{0.225\textwidth}
        \centering
        \includegraphics[width=\linewidth]{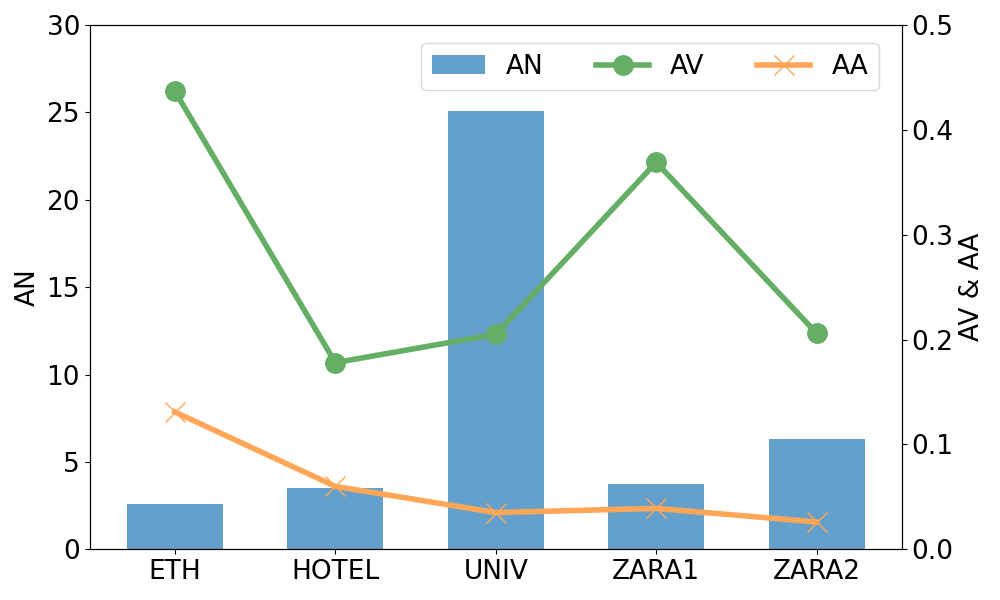} 
        \caption{Density(AN),Speed(AV),and Acceleration(AA) in ETH-UCY.}
        \label{fig:ETH-UCY-statics}
    \end{subfigure}
    \hfill
    \begin{subfigure}{0.22\textwidth}
        \centering
        \includegraphics[width=\linewidth]{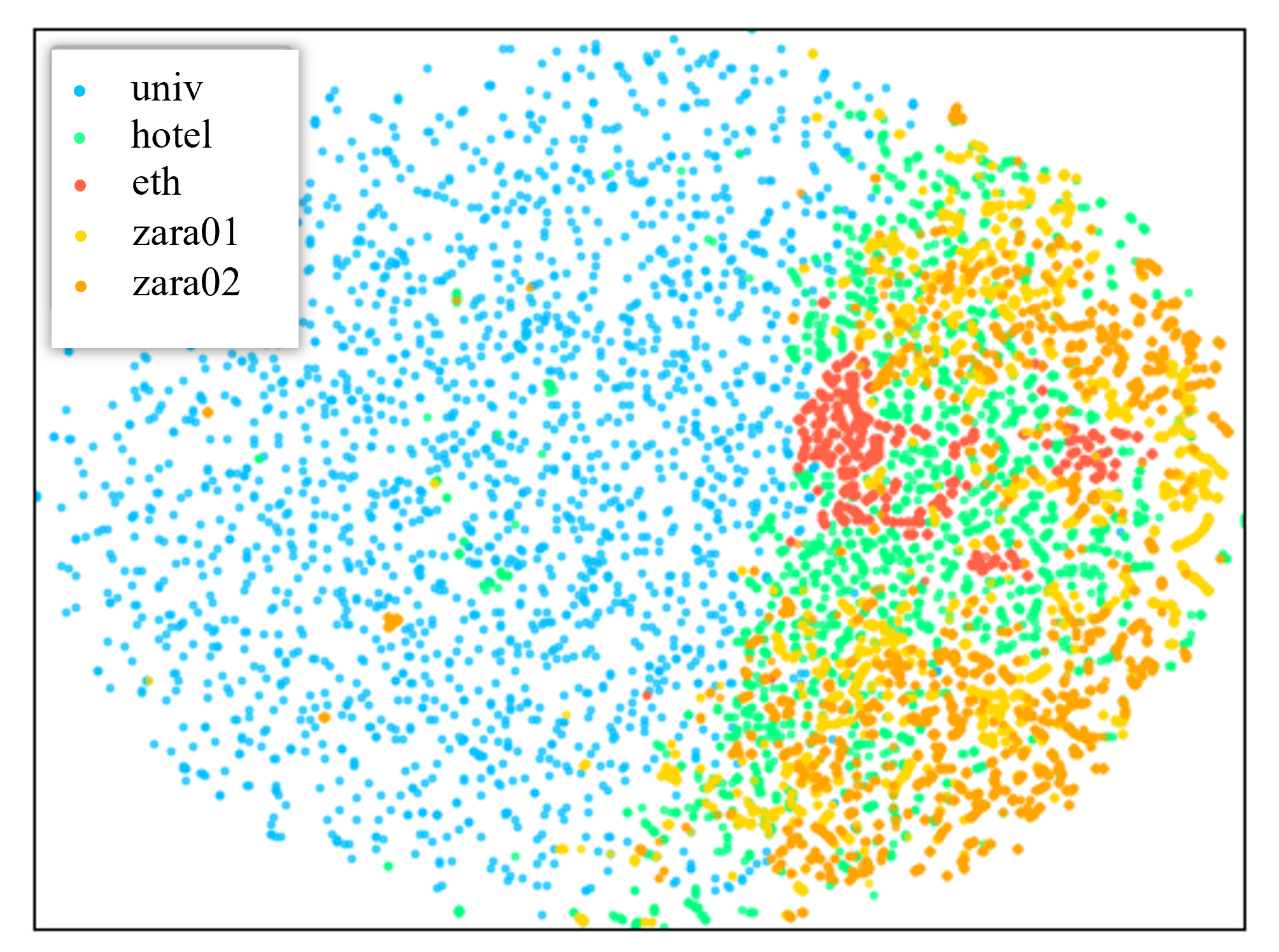}
        \caption{Dual-TT.}
        \label{fig:Dual-TT}
    \end{subfigure}
    \hfill
    \begin{subfigure}{0.22\textwidth}
        \centering
        \includegraphics[width=\linewidth]{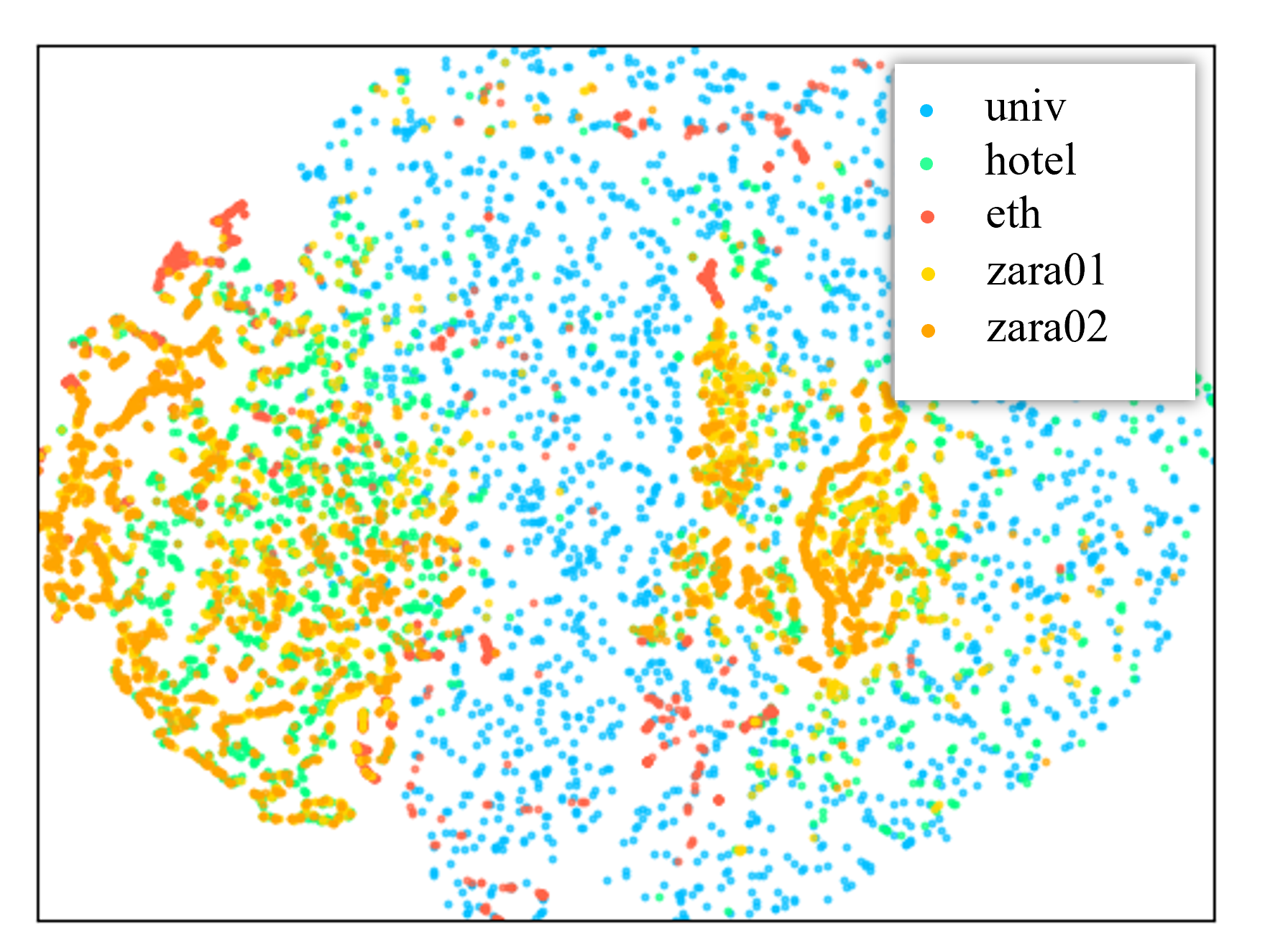}
        \caption{Dual-TT + MGTP.}
        \label{fig:Dual-TT+ MGTP}
    \end{subfigure}
    \hfill
    \begin{subfigure}{0.3\textwidth}
        \centering
        \includegraphics[width=1.2\linewidth]{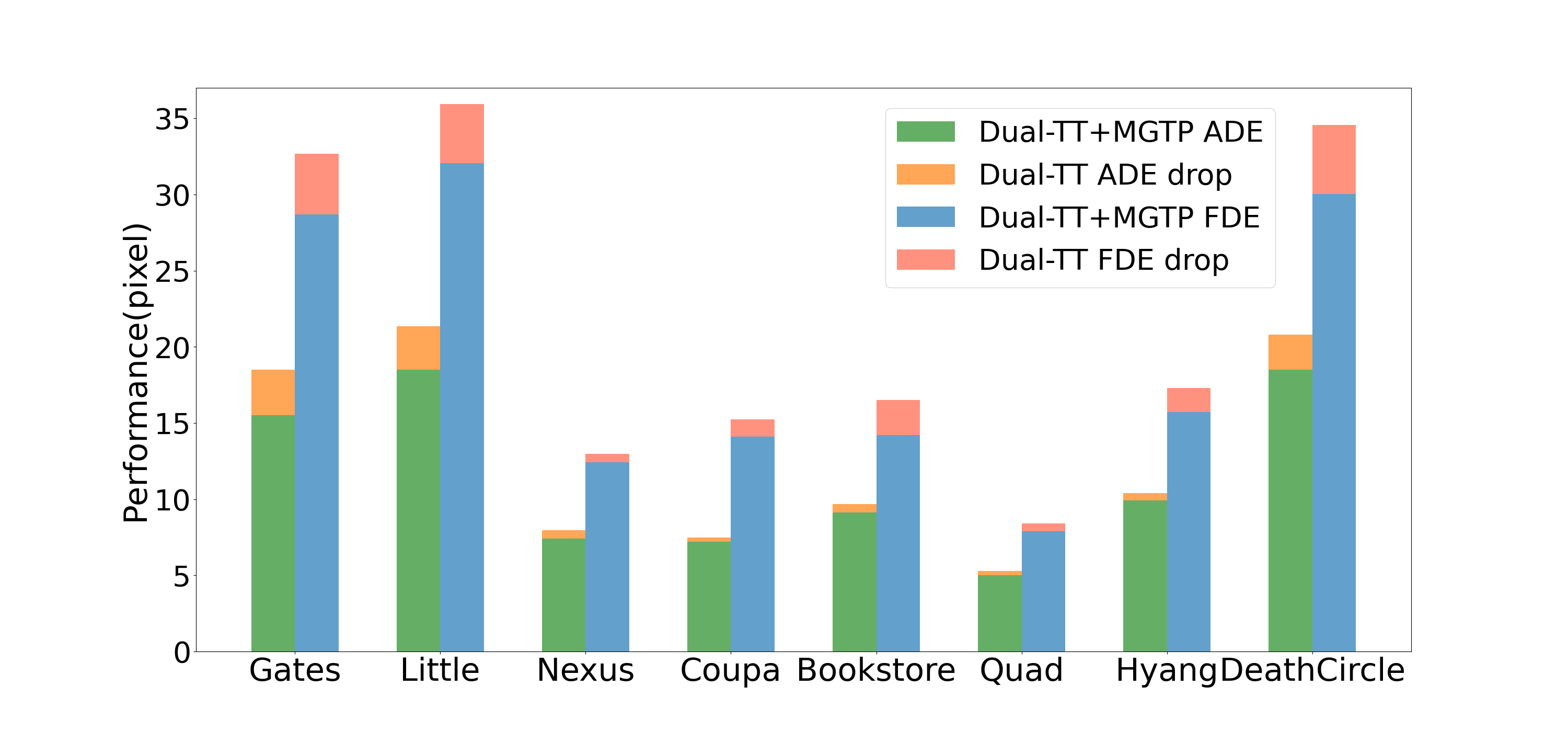}
        \caption{Performance Rise in SDD}
        \label{fig:SDD-split}
    \end{subfigure}
    \caption{Detail analysis for ETH-UCY and SDD.}
    \label{fig:ETH_SDD_detail_analysis_4}
\vspace{-0.5cm}
\end{figure*}
\subsection{Experimental Setup}
\ \indent
We evaluate our method on three real-world public datasets: ETH-UCY \cite{STAR}, SDD \cite{Trajnet} and NBA  \cite{groupnet}.
Each training batch includes 512 agents, spanning various time frames, with training 500 epochs.
The hyperparameters for the serial inner loop learning rate $\lambda$ and parallel outer loop learning rate $\kappa$ are initially set to 0.0015, 0.001. Our method includes 4 parallel optimization paths, each with 4 different serial tasks. We repeat the experiments 5 times and calculate the average to ensure stable results. Details about datasets, baselines, and additional analyses can be found in the Appendix.
\subsection{Prediction Performance}
\ \indent 
\textbf{ETH-UCY Dataset.}
In Table \ref{tab:ETH-UCY}, we compare our approach with several commonly used methods.
To compare with AgentFormer, our method improves the average \text{minADE$_{20}$} from 0.23 to 0.21. Additionally, our method consistently achieves the best or second-best performance across various sub-scenarios.
Particularly in the ETH scenario, where many conventional methods falter, our approach significantly reduces the \text{minADE$_{20}$} from 0.45 to 0.25, representing an efficiency gain of nearly 45\%. By incorporating our MGTP into the AgentFormer model, we achieve a significant performance improvement, resulting in an 8.6\% reduction in \text{minADE$_{20}$} and a 10.2\% reduction in \text{minFDE$_{20}$}.

To investigate the underlying factors contributing to the underperformance of conventional methods in the ETH scenario, 
we collect metrics related to pedestrian density, speed and acceleration. As shown in Fig. \ref{fig:ETH-UCY-statics}, the ETH scenario exhibits the lowest pedestrian density while displaying notably higher average speed and acceleration when compared to other scenarios. This highlights significant disparities between the ETH scenario and the rest.
To further analyze, we employ T-SNE \cite{TSNE} visualization. Without the MGTP framework, as shown in Fig. \ref{fig:Dual-TT}, the ``zara1" and ``zara2" scenarios exhibit overlapping features, indicating similar trajectory designs and spatial distributions. The ``univ" and ``hotel" scenarios have a wider feature distribution, with some interspersed between ``zara1" and ``zara2", while others form distinct clusters. Features in the ETH scenario are distinctly isolated from those in the other four scenarios, leading to discrepancies between decoded trajectories and actual values, resulting in lower model performance. However, with the integration of MGTP framework, as shown in Fig. \ref{fig:Dual-TT+ MGTP}, feature distribution aligns across scenarios, demonstrating improved generalization capabilities.

\textbf{SDD Dataset.}
When applying previous methodologies to this dataset, it results in the creation of two distinct testing configurations. One configuration focuses exclusively on pedestrians, while the other encompasses all types of intelligent agents. To ensure a thorough comparison, we conduct tests under both configurations.
According to our findings presented in Table \ref{tab:SDD}, the integration of the Dual-TT with the MGTP outperforms other methods, resulting in a 12.5\% decrease in the \text{minADE$_{20}$} metric compared to PecNet. 
Additionally, by incorporating the MGTP into the PecNet, we achieve performance improvement, leading to a 10.5\% reduction in \text{minADE$_{20}$} and a 9.44\% reduction in \text{minFDE$_{20}$}.
Futhermore, we adopt a leave-one-out strategy, similar to the ETH-UCY dataset, to revise the partitioning of the SDD dataset. This allows us to conduct additional experiments with Dual-TT model and MGTP framework under new configurations. Our comprehensive findings, shown in Fig. \ref{fig:SDD-split}, consistently demonstrate improved predictive accuracy across all scenarios. Particularly noteworthy are the significant improvements observed in the ``Gates", ``Little", and ``DeathCircle" scenarios, highlighting the effectiveness of the MGTP framework in specific contexts.

\textbf{NBA Dataset.}
\begin{table*}[htpb]
\vspace{-0.2cm}
\setlength{\abovecaptionskip}{0cm}
\setlength{\belowcaptionskip}{-0.2cm}
\centering
\caption{\text{minADE$_{20}$}/\text{minFDE$_{20}$} (meters) results on NBA. GroupNet performs best in previous methods (marked with $*$) and gets better when utilizing MGTP as a plugin. Our Dual-TT, integrated with MGTP, achieves the best results. (bold: best, underline: runner-up)}
\label{tab:NBA}
\resizebox{0.98\textwidth}{!}
{
\begin{tabular}{l|c c c c c |c c|c c}
\hline \hline
\multirow{2}{*}{Time} & Social-STGCNN & Trajectron++ & Agent Former &STAR & PECNet  & GroupNet$^*$ & GroupNet$^*$ & Dual-TT & Dual-TT\\
& \textcolor{blue}{CVPR20\cite{Social-STGCNN}} & \textcolor{blue}{ECCV20\cite{Trajectron++}} & \textcolor{blue}{ICCV21\cite{AgentFormer}} & \textcolor{blue}{ECCV20\cite{STAR}} & \textcolor{blue}{ECCV20\cite{PecNet}} & \textcolor{blue}{CVPR22\cite{groupnet}} & +MGTP &   & +MGTP\\
\hline
1.0s & 0.41/0.62 & 0.30/0.38 & 0.45/0.64 & 0.43/0.66 & 0.40/0.71 & 0.26/0.34 & \underline{0.25/0.32} & 0.26/0.39 & \textbf{0.23/0.35} \\
2.0s & 0.81/1.32 & 0.59/0.82 & 0.84/1.44 & 0.75/1.24 & 0.83/1.61 & 0.49/0.70 & 0.46/0.66 & \underline{0.44/0.71} & \textbf{0.42/0.60} \\
3.0s & 1.19/1.94 & 0.85/1.24 & 1.24/2.18 & 1.03/1.51 & 1.27/2.44 & 0.73/1.02 & 0.70/0.97 & \underline{0.67/1.23} & \textbf{0.64/0.92} \\
4.0s & 1.59/2.41 & 1.15/1.57 & 1.62/2.84 & 1.13/2.01 & 1.69/2.95 & 0.96/1.30 & 0.93/1.21 & \underline{0.87/1.45} & \textbf{0.82/1.37} \\
\hline
\end{tabular}
}
\vspace{0cm}
\end{table*}

Our experimental results, as shown in Table \ref{tab:NBA}, demonstrate that our method consistently outperforms existing techniques across all timestamps in the NBA dataset. 
When compared to the leading baseline method GroupNet, we achieve a significant 14.5\% reduction in \text{minADE$_{20}$}. Furthermore, by integrating the MGTP with GroupNet, we observe respective decreases of 3.7\% and 6.9\% in \text{minADE$_{20}$} and \text{minFDE$_{20}$}.

\begin{figure}[htbp]
\vspace{-0.3cm}
\setlength{\abovecaptionskip}{0.3cm}
\setlength{\belowcaptionskip}{-0.2cm}
    \centering
    \begin{subfigure}{0.23\textwidth}
        \centering
        \includegraphics[width=\linewidth]{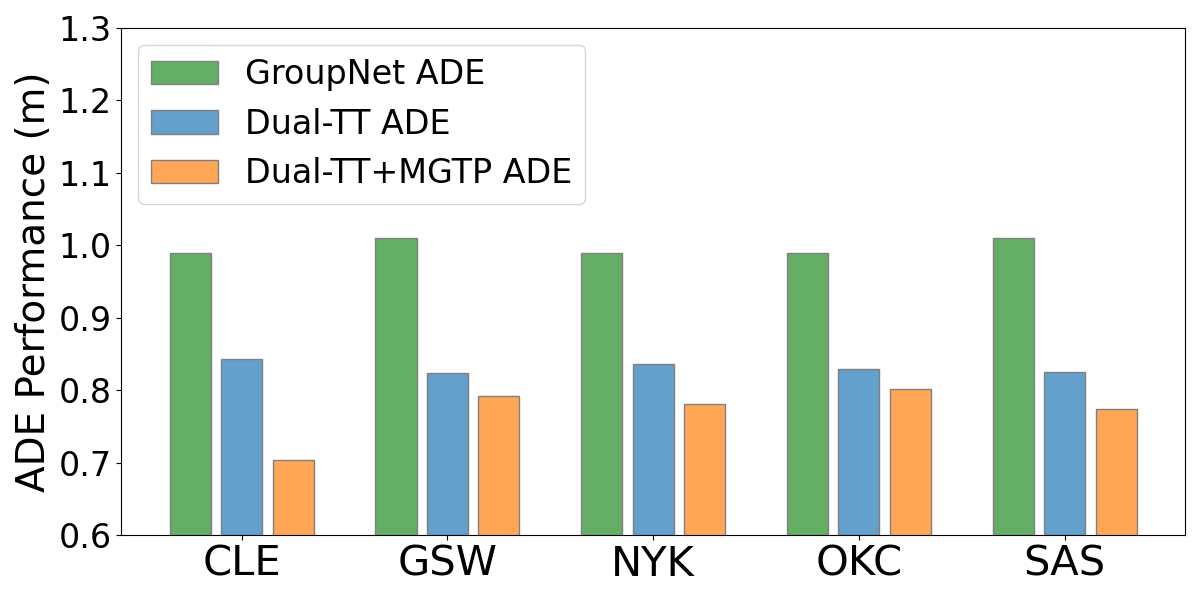}
        \caption{\text{minADE$_{20}$} of 5 NBA teams.}
        \label{fig:nba_ade}
    \end{subfigure}
    \hfill
    \begin{subfigure}{0.23\textwidth}
        \centering
        \includegraphics[width=\linewidth]{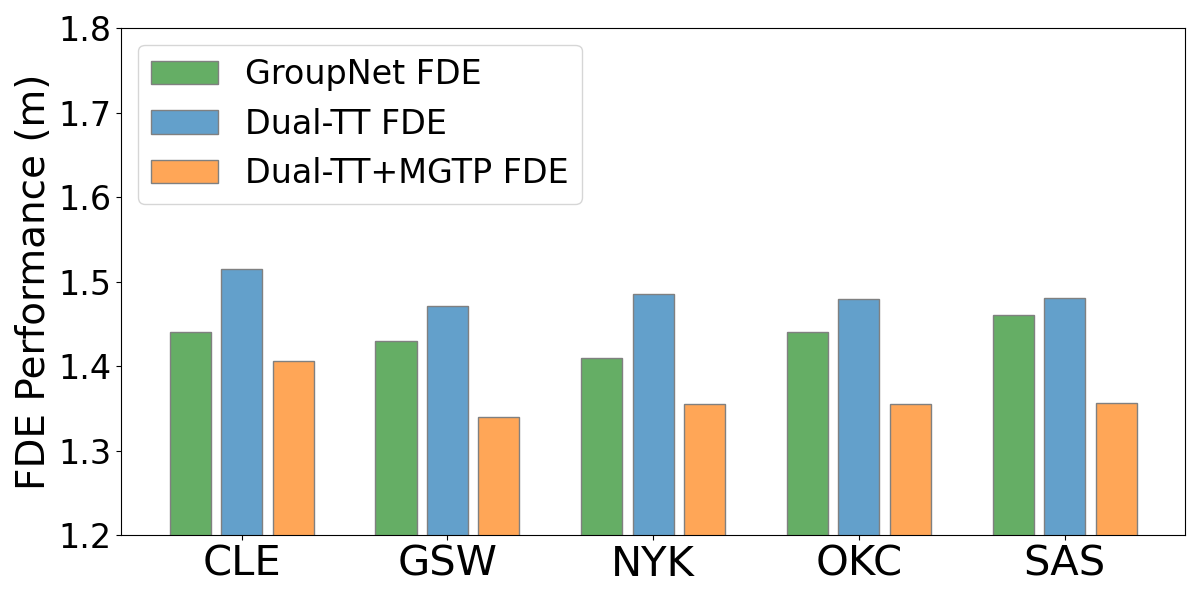}
        \caption{\text{minFDE$_{20}$} of 5 NBA teams.}
        \label{fig:nba_fde}
    \end{subfigure}
    \caption{Performance metrics for 5 NBA teams.}
    \label{fig:nba_performance}
\vspace{-0.5cm}
\end{figure}

We adopt a meta-training and meta-testing partition strategy based on home teams when applying the MGTP framework. This decision is motivated by the consistent geographic scenario observed in the NBA dataset, where athletes exhibit distinct group behavior. Unlike previous pedestrian datasets, the NBA dataset demonstrates cohesive actions among team members, with high interdependencies in their behaviors and trajectories. Notable variations can be observed across different teams, particularly in their offensive and defensive playstyles.
We extract samples from the primary NBA dataset, focusing on teams such as CLE, GSW, NYK, OKC, and SAS. Through leave-one-out experiments, as shown in Fig. \ref{fig:nba_ade} and Fig. \ref{fig:nba_fde}, we consistently observe performance improvements for the Dual-TT model integrated with the MGTP framework when evaluating on these teams as test sets. These results affirm the robustness of the MGTP framework, which organizes the training dataset based on distinct home teams, thereby enhancing model generalization capabilities.

\vspace{-0.1cm}
\subsection{Ablation Studies}
\vspace{-0.1cm}
\ \indent We compare the performance of MGTP with versions that have specific components removed to assess their impact on the model's performance. Our findings are detailed in the subsequent section and Table \ref{tab:Ablation}:
\begin{table*}[htpb]
\setlength{\abovecaptionskip}{0.1cm}
\setlength{\belowcaptionskip}{-0.3cm}
    \centering 
    \caption{\text{minADE$_{20}$}/\text{minFDE$_{20}$} (meters) results of different MGTP variants on the ETH-UCY dataset.}
    \label{tab:Ablation}
    \resizebox{0.75\textwidth}{!}
    {
    \begin{tabular}{c c c |c c c c c |c}
        \hline \hline
         ML & SPT & MM & ETH & HOTEL & UNIV & ZARA1 & ZARA2 & AVG \\
        \hline
        $\times$ & $\times$ & $\times$             & 0.289/0.544 & 0.153/0.281 & 0.315/0.612 & 0.233/0.475 & 0.188/0.378 & 0.2356/0.4580 \\
        $\checkmark$ & $\times$ & $\times$             & 0.279/0.536 & 0.147/0.280 & 0.305/0.606 & 0.213/0.414 & 0.184/0.380 & 0.2263/0.4532 \\
        $\checkmark$ & $\checkmark$ & $\times$     & 0.258/0.491 & 0.136/0.240 & 0.292/0.590 & 0.189/0.396 & 0.169/0.343 & 0.2088/0.4120 \\
        \hline
        $\checkmark$ & $\checkmark$ & $\checkmark$ & 0.245/0.465 & 0.130/0.233 & 0.290/0.586 & 0.184/0.392 & 0.162/0.332 & 0.2034/0.4034 \\
        \hline
    \end{tabular}
    }
\vspace{-0.3cm}
\end{table*}

\textbf{Meta-Learning Basic Module(ML)}:
According to our observations, we note an overall reduction of 8.8\% and 6.8\% in the \text{minADE$_{20}$}/\text{minFDE$_{20}$} metrics, indicating improved performance.
This improvement can be primarily attributed to the combined effect of individual losses during bi-level optimization. 
By aligning gradients in the training phase, this approach ensures consistent gradient updates across domains, leading to enhanced cross-domain performance and reducing overfitting within individual domains.

\textbf{Serial and Parallel Training(SPT)}:
Our approach combines the principles of parallel multi-perspective, which involves combining various optimization paths, and serial multi-tasks, which integrates different domain tasks. This combination serves two purposes: preventing the model from converging to local optima and promoting a thorough exploration of the weight space. 
As shown in Table \ref{tab:Ablation}, our approach consistently outperforms all baselines in the five scenarios evaluated. Particularly notable are the significant increases in \text{minADE$_{20}$} for the ``eth" and ``zara2" scenarios, with improvements of 10.7\% and 11.7\% respectively.

\textbf{MetaMix(MM)}: 
The implementation of the MetaMix strategy has been proven effective in diversifying the meta-testing domain features, resulting in enhanced generalizability of the model. As presented in Table \ref{tab:Ablation}, significant performance improvements can be observed across all scenarios. This highlights the capability of a diverse meta-testing feature set in capturing the complexity and variability of new domains, especially under conditions of domain shift. 
\vspace{-0.1cm}
\subsection{Qualitative Results}
\begin{figure}
\vspace{-0.3cm}
\setlength{\abovecaptionskip}{0.1cm}
\setlength{\belowcaptionskip}{-0.2cm}
    \centering
    \includegraphics[width=0.95\linewidth]{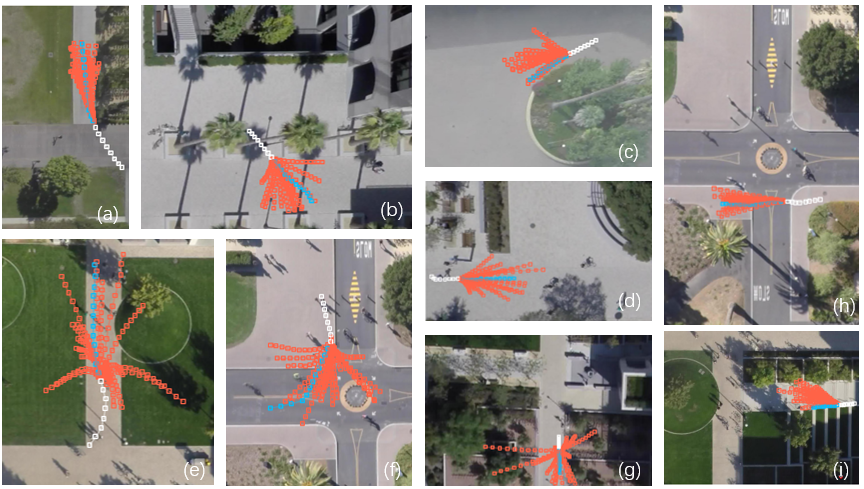}
    \caption{Qualitative examples from SDD: white lines show past trajectories, red lines are predictions from the CVAE decoder, and blue lines represent actual trajectories.}
    \label{fig:SDD-visual-traj}
\vspace{-0.4cm}
\end{figure}

\ \indent Fig. \ref{fig:SDD-visual-traj} presents the multimodal prediction results across various scenarios in the SDD dataset. It is noteworthy that our proposed model's predictions closely align with the ground truth trajectories in the majority of scenarios. Even without relying on map-based context, the model demonstrates proficient navigation around obstacles. For instance, subfigures (a) and (f) show trajectories primarily confined within passable roads, while subfigure (c) exhibits expansion towards obstruction-free areas. In complex scenarios characterized by diverse pedestrian intentions, such as subfigures (e) and (f), the model produces multiple predictions that vary in direction and speed, yet remain plausible. This highlights the effectiveness of our model in generating diverse and realistic predictions in multimodal forecasting tasks.

\vspace{-0.3cm}
\label{sec:experiments}

\section{Conclusion}
\ \indent In this paper, we introduce MetaTra, a novel approach to enhance the generalization performance of trajectory prediction models, focusing on two key aspects: (1) the Dual Trajectory Transformer (Dual-TT), a model adept at assimilating assessments of an individual's intended destination with the dynamics of group motion interactions, and (2) the Meta-Learning for Generalized Trajectory Prediction (MGTP) framework, specifically designed to facilitate trajectory prediction in unseen domain by tuning the training process. Empirical evaluations across several real-world datasets substantiate the efficacy of our approach, showcasing its superior performance in domain generalization. Furthermore, experimental results validate its potential to promote other trajectory prediction models through a plug-and-play unit.
\label{sec:conclusion}

{\small
\bibliographystyle{ieeenat_fullname}
\bibliography{11_references}
}

\ifarxiv \clearpage \appendix \section{MGTP Algorithmic}
The complete learning process of MGTP is summarized in Algorithm \ref{MGTP algorithmic}.
\begin{algorithm}
\caption{Meta-Learning for Generalized Trajectory Prediction (MGTP).}
\begin{algorithmic}[1]
\State \textbf{Input:} source data $D^S = \{D^S_1,...,D^S_N\}$;
\State \textbf{Init:} network parametrized by $\theta$, serial inner loop learning rate $\lambda$, parallel outer loop learning rate $\kappa$, the number of parallel optimization parameter set $C$, the number of serial sampled task $J$;
\While{not converged}
    \For{$c = 1$ to $C$}
        \State $\theta^c_{0} \leftarrow$ Initialized by $\theta$;
        \For{$i = 1$ to $J$}
            \State Randomly sample source tasks $\tau_{S}$ and 
            \State target tasks $\tau_{T}$ within the source domains 
            \State $D^S$;
            \State \textbf{Meta-Train:}
            \State Compute meta-train loss $L_{\tau^i_{S}}(\theta^c_{i-1})$ and 
            \State compute the temporary parameters:
            \State ${\theta'}^c_{i-1} \leftarrow \theta^c_{i-1} - \lambda (L_{\tau^i_{S}}(\theta^c_{i-1}))$;
            \State \textbf{Meta-test:}
            \State Diversify meta-test features with MetaMix;
            \State Compute meta-test loss $L_{\tau^i_{T}}({\theta'}^c_{i-1})$;
        \EndFor
        \State \textbf{Serial Inner Optimization:}
        \State $ \theta^c_{i} \leftarrow \theta^c_{i-1}-\lambda(L_{\tau_{S}^{i}}(\theta^c_{i-1})+L_{\tau^i_{T}}({\theta'}^c_{i-1}))$;
    \EndFor
    \State Get the parallel optimization parameter set 
    \State $\{\widetilde{\theta}_{J}^{1},\ldots,\widetilde{\theta}_{J}^{C}\}$;
    \State \textbf{Parallel Outer Optimization:}
    \State $\theta \leftarrow \theta + \kappa(\sum_{c=1}^{C} \frac{1}{C} \widetilde{\theta}^c_{J} - \theta)$.
\EndWhile
\end{algorithmic}
\label{MGTP algorithmic}
\end{algorithm}
\section{Experiments}
\subsection{Datasets and Preprocessing}
\begin{itemize}

\item\textbf{ETH-UCY Dataset \cite{Trajectron++}:} 
This dataset contains five campus scenarios: ETH, HOTEL, UNIV, ZARA1, and ZARA2, segmenting pedestrian trajectories into 8-second slices. Each slice utilizes an input trajectory of 3.2 seconds (8 frames) to predict a future trajectory spanning 4.8 seconds (12 frames). We select one scenario as the unseen domain for testing by turn while employing the remaining four scenarios for model training, and subsequently calculate the mean of the results.
\item\textbf{StanfordDrone Dataset (SDD) \cite{socialgan}:}
SDD contains approximately 60 videos from 8 primary areas around a university campus, captured via drones. These videos predominantly comprise humans, bikes, and cars. SDD utilizes 3.2 seconds (8 frames) of past data to predict the subsequent 4.8 seconds (12 frames). We follow TrajNet's settings \cite{Trajnet} for fair evaluation, initially focusing on pedestrians as PECNet \cite{PecNet}, then expanding to all agents like Sophie \cite{sophie}. Moreover, each area alternates as the unseen test set, illustrating the effect of MetaTra on generalization.
\item\textbf{NBA SportVU Dataset (NBA) \cite{evolvegraph}:}
This dataset includes trajectories of NBA players and the ball during the 2015-2016 season, collected from SportVU. We follow GroupNet's \cite{groupnet} settings for fair evaluation and utilize 2 seconds (5 frames) to forecast the subsequent 4 seconds (10 frames). Additionally, we categorize data into five home team scenarios (CLE, GSW, SAS, OKC, NYK), and in each test, we randomly select one home team as the unseen domain to achieve domain generalization.
\item\textbf{Soccer Dataset \cite{soccerdataset}:}
The Metrica Sports dataset contains anonymized trackings and event datas, encompassing two games in standard format and one in the EPTS FIFA format. The coordinates in this dataset range from 0 to 1 on each axis, accurately representing the football field of $105x68$ meters. Training is conducted on the NBA dataset, with testing on the Soccer dataset, an analogous athletic context, to validate the model's capability in handling tasks in entirely unseen domains.
\end{itemize}

\subsection{Metrics}
We employ two standard metrics: \text{minADE$_K$} and \text{minFDE$_K$}. \text{minADE$_K$} represents the minimum average distance across all corresponding points between the $K$ predicted trajectories and the actual trajectory. Subsequently, \text{minFDE$_K$} denotes the minimum distance between the final destinations of the $K$ predicted trajectories and that of the actual trajectory.
\subsection{Comparison Methods}
\begin{itemize}
    \item \textbf{STAR \cite{STAR}}: STAR decomposes the spatio-temporal attention modeling into temporal modeling and spatial modeling. It incorporates a memory module and a self-attention approach, enabling sequential frame-by-frame prediction.
    \item \textbf{GroupNet \cite{groupnet}}: GroupNet captures interactions and undertakes representation learning by operating on a multi-scale hypergraph neural network. It integrates CVAE with residual decoders designed for sequence prediction.
    \item \textbf{Social-STGCNN \cite{Social-STGCNN}}: Social-STGCNN employs GCNs to form spatial graphs over time and TCN models for temporal processing.
    \item \textbf{PECNet \cite{PecNet}}:Built upon a social pooling layer, this model employs VAE to estimate potential future endpoints, facilitating subsequent trajectory prediction.
    \item \textbf{Trajectron++ \cite{Trajectron++}}: This model combines a graph-structured framework with a CVAE-based approach, integrating the InfoVAE objective with agent dynamics for enriched trajectory modeling.
    \item \textbf{AgentFormer \cite{AgentFormer}}: Featuring a uniquely designed agent-aware attention mechanism, AgentFormer simultaneously processes temporal and spatial dimensions. It employs CVAE and DLow to generate varied and plausible trajectories.
    \item \textbf{Dual-TT}: We introduce the Dual Trajectory Transformer (Dual-TT) model, which is composed of two pathways: the Interacted-Temporal (IT) and the Temporal-Interacted (TI).
    \item \textbf{MGTP}: We propose the Meta-Learning for Generalized Trajectory Prediction (MGTP) method, which includes an innovative task delineation for the meta-train and meta-test phases. Additionally, it integrates a Serial and Parallel Training (SPT) strategy with the MetaMix method to enhance the stability of the training process.
\end{itemize}

\subsection{NBA-Soccer Transfer}
\begin{table}[htpb]
\vspace{-0.4cm}
    \centering 
    \caption{[NBA $\rightarrow$ Soccer] \text{minADE$_{20}$}/\text{minFDE$_{20}$} (meters) results on NBA-Soccer Transfer settings. Our Dual-TT, integrated with MGTP, achieves the best results. (bold: best, underline: runner-up)}
    \label{tab:NBA-Soccer Transfer}
    \resizebox{0.45\textwidth}{!}
    {
    \begin{tabular}{c| c c c | c c }
        \hline \hline
         metric  & GroupNet & PECNet & STAR & Dual-TT & Dual-TT  \\
                 & \textcolor{blue}{CVPR22\cite{groupnet}} & \textcolor{blue}{ECCV20\cite{PecNet}} & \textcolor{blue}{ECCV20\cite{STAR}} &  & + MGTP \\  
        \hline
        minADE$_{20}$  & 6.983 & 3.310 & 2.337  &  \underline{1.837}  & \textbf{0.966}  \\
        minFDE$_{20}$  & 12.11 & 5.097 & 4.249  &  \underline{3.337}  & \textbf{2.107}  \\
        \hline
    \end{tabular}
    }
\vspace{-0.4cm}
\end{table}
In this experiment, models are trained on a source scene (NBA Dataset) and applied directly to a target scene (Soccer Dataset) without any form of online or offline adaptation. We compare the Dual-TT and Dual-TT+MGTP frameworks against several models known for their strong performance on the NBA dataset. As detailed in Table \ref{tab:NBA-Soccer Transfer}, our methods achieve the best results. Notably, the MGTP significantly boosts the generalization ability of the Dual-TT backbone, yielding improvements of 47.41\% in minADE${_{20}}$ and 36.85\% in minFDE${_{20}}$. 
GroupNet, with its hypergraph structure, outperforms all baselines in the NBA dataset as detailed in Table \ref{tab:NBA} of the main body. However, it exhibits limited transferability across different datasets due to its reliance on dataset-specific hyperparameter tuning. PECNet, which predicts trajectories through endpoint estimation, also shows subpar performance. Meanwhile, STAR, another model considering temporal-spatial factors, performs even worse than Dual-TT. In contrast, our Dual-TT demonstrates superior adaptability across diverse datasets, not dependent on highly specialized hyperparameters. Consequently, MGTP demonstrates robust generalization capabilities in this context due to its reduced dependency on the dataset's domain.

\subsection{Parameter Sensitivity}
Within the SPT framework, we investigate the impact of two pivotal parameters on model performance: the number of serial inner updates $J$ and parallel outer optimizations $C$. As shown in Fig. \ref{fig:sdd-trajectories}, our results support the theory that a varied set of serial tasks improves the model's ability to adapt to new situations, but only to a certain extent. Preceding tasks are crucial for an excellent start to learning while adding more tasks later has limited impact. Fig. \ref{fig:sdd-tasks} demonstrates that utilizing four parallel outer optimizations results in stable and effective outcomes. However, increasing the number of parallel optimizations beyond this threshold yields marginal benefits while incurring higher computational costs.
\begin{figure}[]
\vspace{-0.5cm}
\setlength{\abovecaptionskip}{0.3cm}
\setlength{\belowcaptionskip}{-0.2cm}
    \centering
    \begin{subfigure}{0.23\textwidth}
        \centering
        \includegraphics[width=\linewidth]{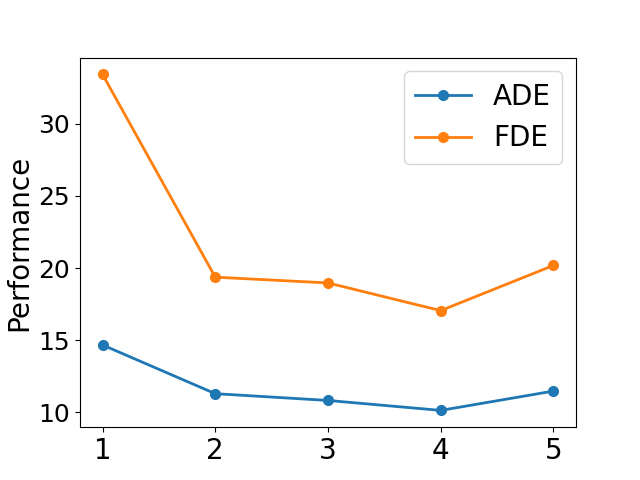}
        \caption{The number of serial inner updates $J$.}
        \label{fig:sdd-trajectories}
    \end{subfigure}
    \hfill
    \begin{subfigure}{0.23\textwidth}
        \centering
        \includegraphics[width=\linewidth]{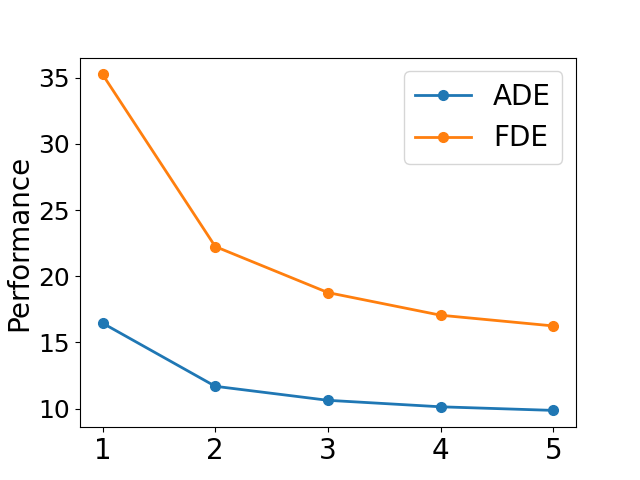}
        \caption{The number of parallel outer optimizations $C$.}
        \label{fig:sdd-tasks}
    \end{subfigure}
    \caption{The impact of parameters $J$ and $C$ in SPT.}
    \label{fig:Parameter sensitivity of T and S on SDD}
\end{figure}
\subsection{More Qualitative Results}
As illustrated in Fig. \ref{fig:comparison-SDD}, we showcase the visually optimal prediction outcomes from several models within a diverse set of unseen test scenarios from the SDD dataset. Our model's results are most proximate to the ground truth, exhibiting the smallest distances at corresponding points. 
In Fig. \ref{fig:comparison-SDD} (a), within the ``Coupa" scenario, the Agent-Former deviates from the ground truth.
In Fig. \ref{fig:comparison-SDD} (b)(c)(d), models such as PecNet, GroupNet, and Agent-Former fail to accurately predict the agent's trajectory along the road, instead veering off the designated path. This could imply that in scenarios involving turns, circumnavigating obstacles, or navigating complex environments, the predictive performance of these models falls short compared to our model.
\begin{figure}[htbp]
\vspace{-0.4cm}
\setlength{\abovecaptionskip}{0.3cm}
\setlength{\belowcaptionskip}{-0.2cm}
    \centering
    \includegraphics[width=0.95\linewidth]{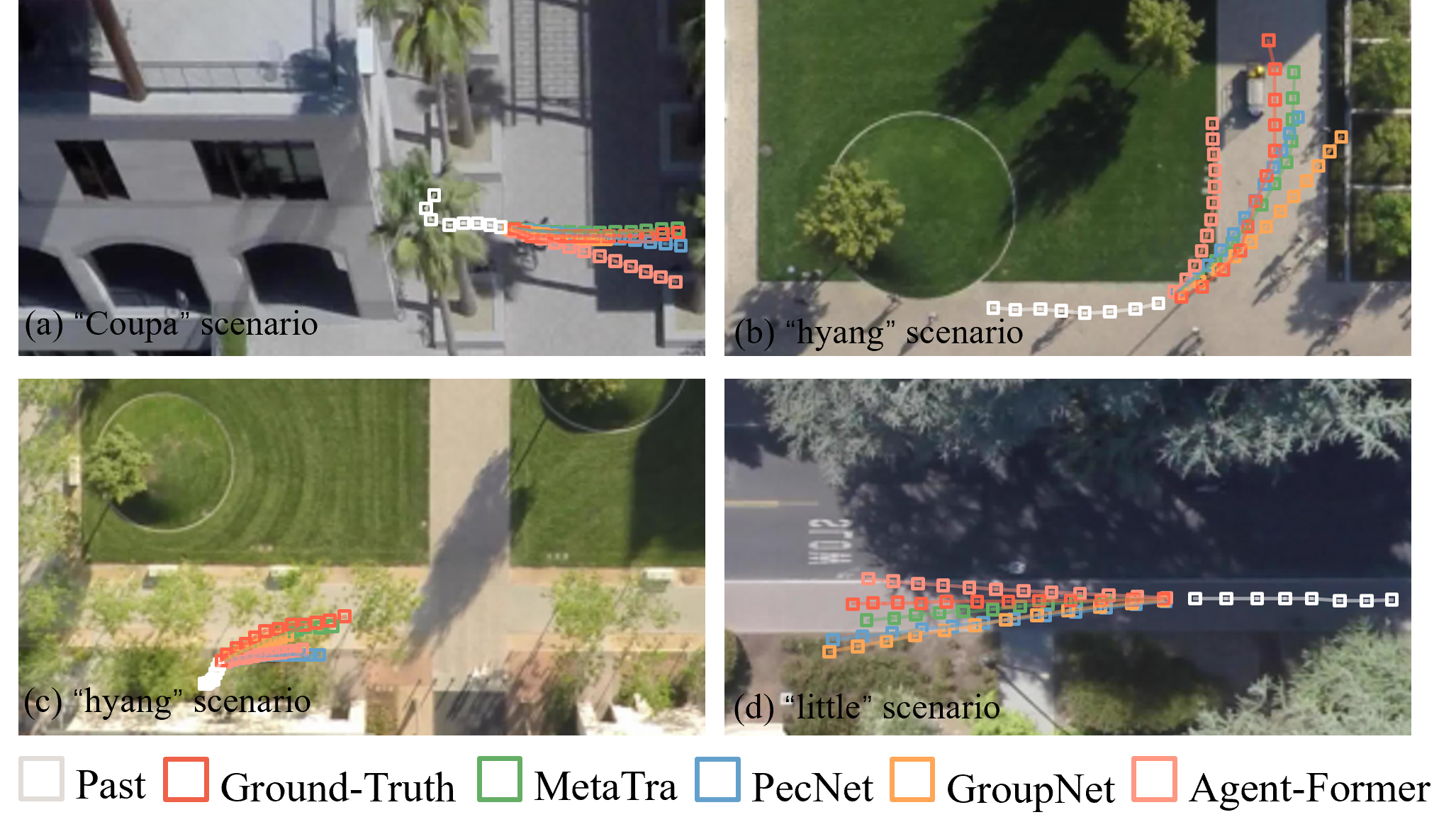}
    \caption{Qualitative comparative analysis of different methods on SDD.}
    \label{fig:comparison-SDD}
\end{figure}

 \fi

\end{document}